\documentclass{article}
\usepackage{ijcai19}

\usepackage{times}
\usepackage{soul}
\usepackage{url}
\usepackage[utf8]{inputenc}
\usepackage[small]{caption}
\usepackage{graphicx}
\usepackage{amsmath}
\usepackage{booktabs}
\usepackage{algorithm}
\usepackage{algorithmic}
\usepackage{epstopdf}
\usepackage{float}
\usepackage{textcomp}
\usepackage{balance}

\usepackage[usenames,dvipsnames]{color}

\usepackage[normalem]{ulem}
\usepackage{multirow}
\usepackage{array}
\usepackage{amsmath}
\usepackage{amssymb}
\usepackage{bm}

\DeclareMathOperator*{\argmax}{arg\,max}

\title{Improving Interactive Reinforcement Agent Planning with Human Demonstration}

\author{
        Guangliang Li$^{1}$\and
        Randy Gomez$^{2}$\and
        Keisuke Nakamura$^{2}$\and
        Jinying Lin$^{1}$\and
        Qilei Zhang$^{1}$\And
        Bo He$^{1}$
    \affiliations
    $^{1}$Department of Electronic Engineering, Ocean University of China, China\\ 
    $^{2}$Honda Research Institute Japan Co. Ltd., Japan
    \emails
    \{guangliangli, bhe\}@ouc.edu.cn, 
    \{r.gomez, k.nakamura\}@jp.honda-ri.com/.
}

\begin{document}

\maketitle

\begin{abstract}
TAMER has proven to be a powerful interactive reinforcement learning method for allowing ordinary people to teach and personalize autonomous agents' behavior by providing evaluative feedback. However, a TAMER agent planning with UCT---a Monte Carlo Tree Search strategy, can only update states along its path and might induce high learning cost especially for a physical robot. In this paper, we propose to drive the agent's exploration along the optimal path and reduce the learning cost by initializing the agent's reward function via inverse reinforcement learning from demonstration. We test our proposed method in the RL benchmark domain---Grid World---with different discounts on human reward. Our results show that learning from demonstration can allow a TAMER agent to learn a roughly optimal policy up to the deepest search and encourage the agent to explore along the optimal path. In addition, we find that learning from demonstration can improve the learning efficiency by reducing total feedback, the number of incorrect actions and increasing the ratio of correct actions to obtain an optimal policy, allowing a TAMER agent to converge faster. 


\end{abstract}

\section{Introduction}

Autonomous agents have the potential to operate in most applications in the human's living environment in the near future. In the real-world, as the increasing interactions between people and agents, users may want to teach agents an optimal behavior and even customize agents' behavior according to their preferences. Interactive reinforcement learning \cite{thomaz2008teachable,knox2009interactively,loftin2015learning,macglashan2017interactive} 
has been developed and proved to be a powerful method for facilitating non-technical people to teach an agent to perform a task 
using evaluations of the quality of the agent's behavior. 

In this work, we focus on interactive reinforcement agent planning and extend the TAMER framework \cite{knox2009interactively}---a popular interactive reinforcement learning method. 
To facilitate a TAMER agent to plan in complex environments, TAMER was combined with Upper Confidence Bounds for Trees (UCT) \cite{kocsis2006bandit}, a typical Monte Carlo Tree Search strategy \cite{knox2015framing}. However, the TAMER agent planning with UCT can only update states along its path, which causes a local-bias problem and results in low learning efficiency. For example, when the agent does not visit the states along the optimal path, the human trainer cannot give feedback for those states, preventing the agent from learning about what reward might be received along those critical states. Therefore a TAMER agent planning with UCT ironically needs to traverse to the goal to learn the accurate reward predictions that might lead it to traverse to the goal \cite{knox2015framing}. 
In addition, when learning from human-generated reward, a TAMER agent still learns from trial-and-error. 
In some situation, this will make the agent learning dangerous or induce high cost, especially for the physical robot learning, e.g., learning to drive a car. 

In this paper, we try to 
solve the encountered local-bias problem when the TAMER agent plans with UCT 
and reduce the agent's cost in the learning process. We propose to improve the TAMER agent's exploration along the optimal path by initializing the agent's reward function via inverse reinforcement learning from human demonstration, which is another main natural teaching methods developed for enabling autonomous agents to learn from a non-technical teacher \cite{argall2009survey}. Learning from demonstration will often lead to faster learning than reward signals and highlight a subspace for the agent to explore.
We evaluate our proposed method in a RL benchmark domain---Grid World---with different discounts on human reward. This is the first time that a TAMER agent planning with UCT has been tested with real human user and with different discounts on human reward. Our results indicate the usefulness of our proposed method on solving the problem of overly local updates and reducing the learning cost for a TAMER agent planning with UCT. 

\section{Related Work}
\label{sec:rw}




When learning from human reward, an agent uses the evaluations of its behavior provided by a human trainer to improve its behavior \cite{isbell2001social,knox2009interactively,pilarski2011online,suay2011effect}. 
Thomaz and Breazeal \cite{thomaz2008teachable} implemented an interface with a tabular \emph{Q-learning} \cite{watkins1992q} agent where a separate interaction channel was provided allowing the human to give the agent feedback. The agent aims to maximize its total discounted reward, which is the sum of human reward and environmental reward. 
Suay and Chernova \cite{suay2011effect} extended their work to a real-world robotic system using only human reward. Knox and Stone \cite{knox2009interactively} proposed the \emph{TAMER} framework that allows an agent to learn from only human reward signals instead of environmental rewards by directly modeling the human reward. Moreover, 
Warnell et al. \cite{warnell2017deep} proposed Deep TAMER, an extension of the TAMER framework that leverages the representational power of deep neural networks in order to learn complex tasks with high-dimensional state spaces. In addition, MacGlashan et al. \cite{macglashan2017interactive} claimed human evaluative feedback to be interpreted as policy feedback depending on the agent's current policy and proposed an actor-critic algorithm--Convergent Actor-Critic by Humans (
COACH
), to learn from human feedback. 
Loftin et al.\ \cite{loftin2014strategy} 
interpreted human feedback as categorical feedback strategies that depend both on the behavior the trainer is trying to teach and the trainer's teaching strategy.

In learning from demonstration, the agent learns from sequences of state-action pairs provided by a human trainer who demonstrates the desired behavior \cite{argall2009survey}. For example, \emph{apprenticeship learning} \cite{abbeel2004apprenticeship} is a form of learning from demonstration, which learns how to perform a task using \emph{inverse reinforcement learning} \cite{ng2000algorithms} from observations of the behavior demonstrated by an expert teacher. 
Argall et al. \cite{argall2007learning} proposed a method wherein the agent learns from both demonstrations and the trainer's critiques of the agent's task performance, which is quite related to our work in this paper. However, our work differs in allowing the human trainer to provide human rewards --- evaluations of the quality of the agent's action --- to fine-tune the agent's behavior, while in their work only the critiques of the whole task's performance were provided.  


The work of Judah et al.\ \cite{judah2014imitation} is most related to our work in this paper. Specifically, they used a specified shaping reward function to improve the learning efficiency of learning from demonstration. However, our work differs by allowing the shaping reward to be provided by a human trainer not pre-defined potential function by the agent designer. In addition, Brys et al.\ \cite{brys2015reinforcement} proposed a method for speeding up reinforcement agent learning from environmental rewards by reward shaping via a learned potential function from demonstrations. While in our work, we used demonstrations to seed the agent's learning from human-generated rewards. 

\section{Preliminaries: Interactive Reinforcement Learning}
\label{sec:pre}



Interactive reinforcement learning (Interactive RL) was developed to allow  
an ordinary human user 
to shape the agent learner by providing evaluative feedback \cite{thomaz2008teachable,knox2009interactively,
loftin2015learning,macglashan2017interactive}. 

As in traditional reinforcement learning (RL) \cite{sutton1998reinforcement}, 
an interactive RL agent learns to make sequential decisions in a task. 
A sequential decision task is modeled as a Markov decision process (MDP), denoted by a tuple \{$S$, $A$, $T$, $R$, $\gamma$\}. In MDP, time is divided into discrete time steps, and $S$ is a set of states in the environment. 
and $A$ is a set of actions that the agent can perform. At each time step $t$, the agent observes the state of the environment, $s_{t}$ $\in$ $S$ 
and takes an action $a_{t}$ $\in$ $A$. The experienced state-action pair will take the agent into a new state $s_{t+1}$, 
decided by a transition function 
$T(s_{t},a_{t},s_{t+1}) = Pr(s_{t+1}|s_{t}, a_{t})$.
The agent will receive an evaluative feedback $h_{t+1}$, provided by a human observer 
evaluating the quality of the action selection based on her knowledge. That is to say, there is no predefined reward function in interactive RL. 
The discount factor $\gamma$ 
determines the present value of future rewards. 
The objective of the agent is to learn a policy $\pi$, mapping from states to actions. 
One common way is to learn an action value function, $Q(s, a)$
, which estimates the long-term discounted reward for a given state-action pair. Given an optimal action value function, the optimal policy can be obtained by greedily selecting the action with the highest value in the current state.

\subsection{The TAMER Framework}
\label{sec:tamer}

\begin{figure} [htb]
\centering
\vspace{-0.1cm}
\includegraphics[width=2.8in]{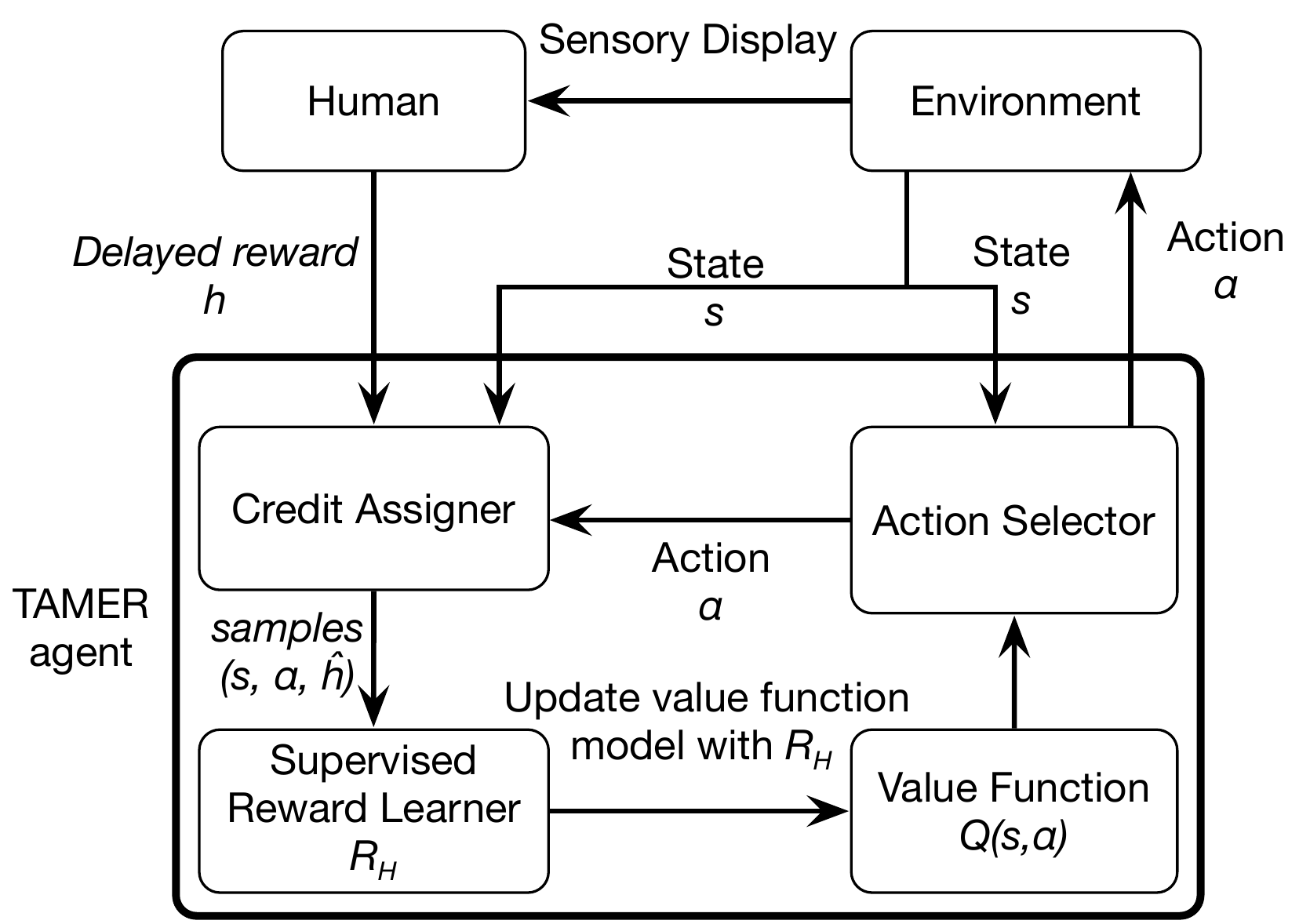}
\vspace{-0.1cm}
\caption{An agent learns from a human teacher with TAMER (modified from \protect\cite{knox2009interactively}).}
\label{tamer}
\vspace{-0.2cm}
\end{figure}

In this paper, we use the TAMER framework \cite{knox2009interactively} as the agent's learning algorithm. TAMER is a typical interactive reinforcement learning method. Different from the original TAMER framework which learns and selects actions with the reward function \cite{knox2009interactively}, in this paper, we rephrased TAMER as a general model-based method for agent learning from human reward, as shown in Figure \ref{tamer}.

An agent implemented according to TAMER learns from real-time evaluations of its behavior, provided by a human teacher who observes the agent. These evaluations are taken as human reward signals. The TAMER agent learns a model of the human reward and then uses it 
to learn a value function. The TAMER agent will select actions with the value function to get the most accumulated human reward.
 
There are four key modules for an agent learning with TAMER. The first one is to learn a predictive model of human reward. Specifically, the TAMER agent learns a function $\hat{R}_{H}(s, a): S \times A \rightarrow \Re$, approximating the expectation of human rewards received in the interaction experience:
\begin{equation}
 \hat{R}_{H}(s, a) = \vec{w}^{\mathrm{T}}\Phi(s,a), 
\end{equation}
where $\vec{w}$ 
is the parameter vector, and $\Phi(\vec{x})$ 
is a vector of basis feature functions. 

The second one is the credit assigner to deal with the time delay of human reward caused by evaluation of agent's behavior and delivering it. 
TAMER defines a probability density function $f(t)$ to estimate the probability of the teacher's feedback delay. 
$f(t)$ provides the probability that the feedback occurs within any specific time interval and is used to calculate the probability (i.e. the credit) that a single reward signal is targeting a single time step. At the current time step $t$, the credit for each previous time step $t$-$k$ is computed as:

\begin{equation}
c_{t-k} = \int_{t-k-1}^{t-k} f(x) dx.
\label{credit}
\end{equation}

If a human teacher gives multiple rewards, the label $\hat{h}$ for each previous time step (state-action pair) is the sum of all credits calculated with each human reward using Equation \ref{credit}. 
The TAMER agent uses $\hat{h}$ 
and state-action pair as a supervised learning sample to learn 
$\hat{R}_{H}(s, a)$ by updating its parameters, e.g., with incremental gradient descent: 

\begin{equation}
\begin{aligned}
\delta_{t} 
&= \hat{h} - \vec{w}^{\mathrm{T}}\Phi(s_{t},a_{t}),
\end{aligned}
\label{herror}
\end{equation}

\vspace{-0.4cm}

\begin{equation}
\begin{aligned}
\vec{w}_{t+1} 
&= \vec{w}_{t} + \alpha \delta_{t}\Phi(s_{t},a_{t}),
\end{aligned}
\label{rupdate}
\end{equation}
where $\alpha$ is the learning rate,  $\delta_{t}$ is temporal difference error.

The third one is the value function module. The TAMER agent learns 
an action value function---$Q(s,a)$ from the learned human reward function $\hat{R}_{H}(s, a)$: 
\begin{equation}
Q(s,a) \leftarrow \hat{R}_{H}(s, a)+\gamma \sum_{s' \in S}T(s,a,s') \times max_{a'}Q(s',a'), 
\end{equation}

%


The fourth module is the action selector. 
As a traditional RL agent which seeks the largest discounted accumulated future rewards, a TAMER agent also greedily selects the action with the largest value: 
%


\begin{equation}
a \leftarrow \argmax_{a}[\hat{R}_{H}(s, a)+\gamma \sum_{s' \in S}T(s,a,s') \times max_{a'}Q(s',a'))]. 
\end{equation}

The TAMER agent learns by repeatedly taking an action, sensing reward, and updating the predictive model $\hat{R}_{H}$ and corresponding value function. 

\subsection{Inverse Reinforcement Learning}
\label{sec:efa}

Similar to TAMER, in inverse reinforcement learning (IRL), an agent also learns in an MDP$\backslash$R \cite{ng2000algorithms}. 
An agent learning via IRL 
assumes that there is a vector of features $\phi$ defined over states, and a ``true" unknown reward function $R^{\ast} = w^{\ast} \cdot \phi(s)$ on which the demonstrator is trying to optimize \cite{abbeel2004apprenticeship}, where $\phi(s)$ is the vector of basis functions and $w$ is the weight vector. The value of a policy $\pi$ is calculated as

\begin{equation}
\begin{aligned}
V(\pi)  & = E[ \Sigma_{k=0}^{\infty}\gamma^{k}R(s_{t}) |\pi] \\
          & = w \cdot E[ \Sigma_{k=0}^{\infty}\gamma^{k} \phi(s_{t})) |\pi],
\end{aligned}          
\end{equation}
where $V(\pi)$ is the state value function following policy $\pi$, $\gamma$ is the discount factor, $\phi(s_{t})$ is a vector of basis functions describing features over state $s_{t}$, $w$ is the weight vector over the basis functions for the reward function $R$.

The feature expectations $\mu(\pi)$ is defined as the expected discounted accumulated feature value vector, calculated as: 
\begin{equation}
\mu(\pi) = E[ \Sigma_{k=0}^{\infty}\gamma^{k} \phi(s_{t})) |\pi].
\end{equation}
With this notation, the value of a policy $\pi$ can be written as 
\begin{equation}
\label{eq:value}
V(\pi)  = w \cdot \mu(\pi).
\end{equation}
Therefore, if we can find the optimal (or close to optimal) weight vector $w$ for the reward function $R$, then the optimal value function of policy $\pi$ can be derived with Equation \ref{eq:value}, which can attain performance near that of the demonstrator on the unknown reward function $R$. In this paper, we use the projection algorithm \cite{abbeel2004apprenticeship} in our proposed method.

\section{Methodology}
\label{sec:md}

The TAMER agent learning can be combined with planning methods such as value iteration, Monte Carlo Tree Search etc. In this case, the agent 
uses simulated experience to speed up its learning. Value iteration updates the state value for the whole state space. However, in considerably more complex domains which usually have large state space, agents will not be able to perform value iteration by sweeping over the entire state space, let alone those tasks with continuous states and actions \cite{knox2015framing}.

TAMER can also update the value function through a Monte Carlo Tree Search (MCTS) strategy---Upper Confidence Trees (UCT) \cite{knox2015framing}. These MCTS-based search has been successfully applied in many complex tasks with especially large state spaces \cite{kocsis2006bandit}, e.g., the game of Go \cite{silver2017mastering}. 
However, the TAMER agent planning with UCT can only update states along its path, which causes a local-bias problem and results in low learning efficiency. Moreover, when the agent does not visit the states near the goal, the trainer cannot give feedback for those states, preventing the agent from learning about what reward might be received along those critical states \cite{knox2015framing}. 




Optimistically initializing the reward function might solve the local-bias problem since it can drive the agent to explore the whole state-action space \cite{knox2015framing}. However, this will lead to extensive exploration, 
and might frustrate, exhaust, and even confuse the human trainer, since with such thorough exploration the agent may not respond to the trainer's feedback and be unable to learn for a considerable period of time. Moreover, this 
would additionally sacrifice the fast learning 
which is the main advantage of interactive reinforcement learning. Nonetheless, we believe some mild optimistic initialization might solve the problem of overly local exploration. 


Therefore, to drive the agent's exploration 
and improve the agent's learning efficiency, in this paper, we propose to use human demonstration to seed the TAMER agent learning and planning. Specifically, our proposed method allows the agent to learn a reward function from human provided demonstrations via inverse reinforcement learning (IRL) first. The demonstrations provided by the human trainer typically consist of sequences of state-action pairs \{$(s_{0}, a_{0}), ..., (s_{n}, a_{n})$\}, which will be fed into the IRL algorithm. The learned reward function via IRL from the demonstration is used to seed the reward function $\hat{R}_{H}$ in TAMER, which learns a value function with learned human reward function and plans with UCT at the same time. Then the human trainer can revise the agent's policy with human rewards. 

In our approach, inverse reinforcement learning was implemented with the projection algorithm \cite{abbeel2004apprenticeship}, though approaches such as maximum entropy, bayesian and game-theoretic can also be used. In IRL, we implemented UCT to generate planning trajectories for optimizing the reward function. 
In TAMER, planning trajectories are also chosen by UCT, where 
the search tree is reset at the start of each time step to respect the corresponding change to the human reward function $\hat{R}_{H}$ in TAMER, which generally makes past search results inaccurate. 

In this paper, we would like to test and see how 
human demonstration will improve a TAMER agent's learning and drive its planning from human reward, not to solve the task with demonstrations or human reward alone. Therefore, as a starting point, we assume the human trainer prefer to provide one demonstration first and then use human reward to revise the agent's behavior, though more demonstrations can be provided even until the problem in the task is solved with only demonstrations. However, we will investigate the effect of more demonstrations on agent's learning and planning from human reward and even the interchangeability of demonstrations and human rewards in future work.

\section{Experiments}
\label{sec:ep} 

\begin{figure} [htb]
\centering
\includegraphics[width=1.7in]{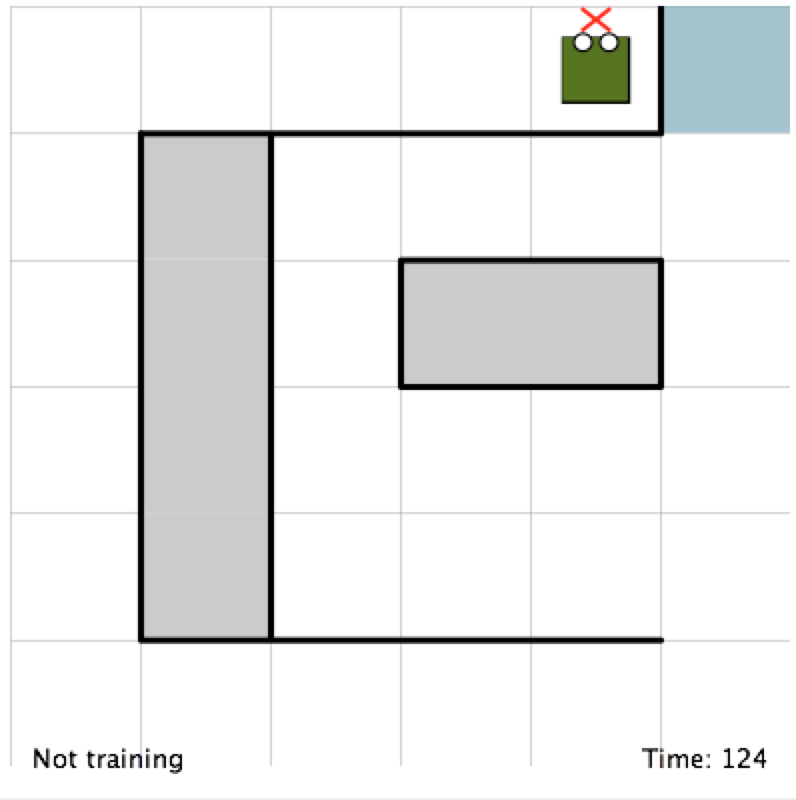}
\caption{A screenshot of the Grid World domain. The robot's current location is the starting state for each episode, and the goal state is the blue block next to it with a wall between them. Note that the dark black lines and the grey blocks are walls.}
\label{gridworld}
\vspace{-0.2cm}
\end{figure}

To demonstrate the potential usefulness of our proposed approach, in this paper, we perform experiments in the Grid World domain---a benchmarking problem in reinforcement learning. 
The grid world task contains 30 states. For each state, at each time step the agent can choose from four actions: moving up, down, left or right. The action attempted through a wall results in no movement for that step. Task performance metrics are based on the number of time steps (actions) taken to reach the goal. The agent always starts one learning episode in the same state, which is shown as the robot's location in Figure \ref{gridworld}. The red cross indicates the direction of the agent's action. In the task, the agent tries to learn a policy that can reach the goal state (the blue square in Figure \ref{gridworld}) with as few time steps as possible. The optimal policy from the start state requires 19 actions.

In our experiments, to see the effect of our proposed method, 
we compare the agent learning via TAMER to that via our proposed method with different discount rates on human reward ($\gamma_{UCT}$). 
The TAMER framework and the TAMER module in our proposed method are the same and both search with the UCT algorithm to update the action value function. Therefore, when we say $\gamma_{UCT}$, it applies to both of them. The only difference between TAMER and our method is whether learning from demonstration via IRL 
is incorporated or not. In addition, both reward functions $R_{H}$ in TAMER and $R$ in IRL, action value functions $Q(s,a)$ in TAMER and IRL, are represented by a linear model of Gaussian radial basis functions. 

One radial basis function is centered on each cell of the grid world, effectively creating a pseudo-tabular representation that generalizes slightly between nearby cells. Each radial basis function has a width $\sigma^{2}=0.05$, where 1 is the distance to the nearest adjacent center of a radial basis function, and the linear model has an additional bias feature of constant value 0.1 \cite{knox2015framing}. The discount factor $\gamma$ for learning from demonstration via IRL is set to 0.99 ($\gamma_{IRL}$). The author of the paper trained both agents learning via our proposed method and TAMER with all discount factors. Each agent with every discount factor was trained for 10 trials. For each trial with either method, we trained the agent to the best ability until an optimal policy is obtained. 
With our proposed method, we first provide one single demonstration navigating from the start state to the goal state via keyboard, and then train the agent with human rewards as in TAMER. Note that, for each trial, the demonstration was performed to the best ability but might not always be optimal. The analysis in the next section is based on an average of data collected from the 10 trials. 

\section{Experimental Results}
\label{sec:re}

This section provides results over experiments performed in the Grid World domain with our proposed algorithm in comparison with TAMER. 
Our experiments were conducted with discount factor $\gamma_{IRL}$ = 0.99 for learning from demonstration via IRL, paired with discount factor $\gamma_{UCT}$ = 0, 0.7, 0.9 and 0.99 on human reward for TAMER agent planning with UCT. Note that $\gamma_{UCT}$ for the TAMER module in our proposed algorithm 
and TAMER 
are with the same values.

\subsection{Number of Feedback}

\begin{figure}[htb]
\centering
\begin{tabular}{c c}
\includegraphics[width=0.48\columnwidth,  trim=40 0 0 0, clip]{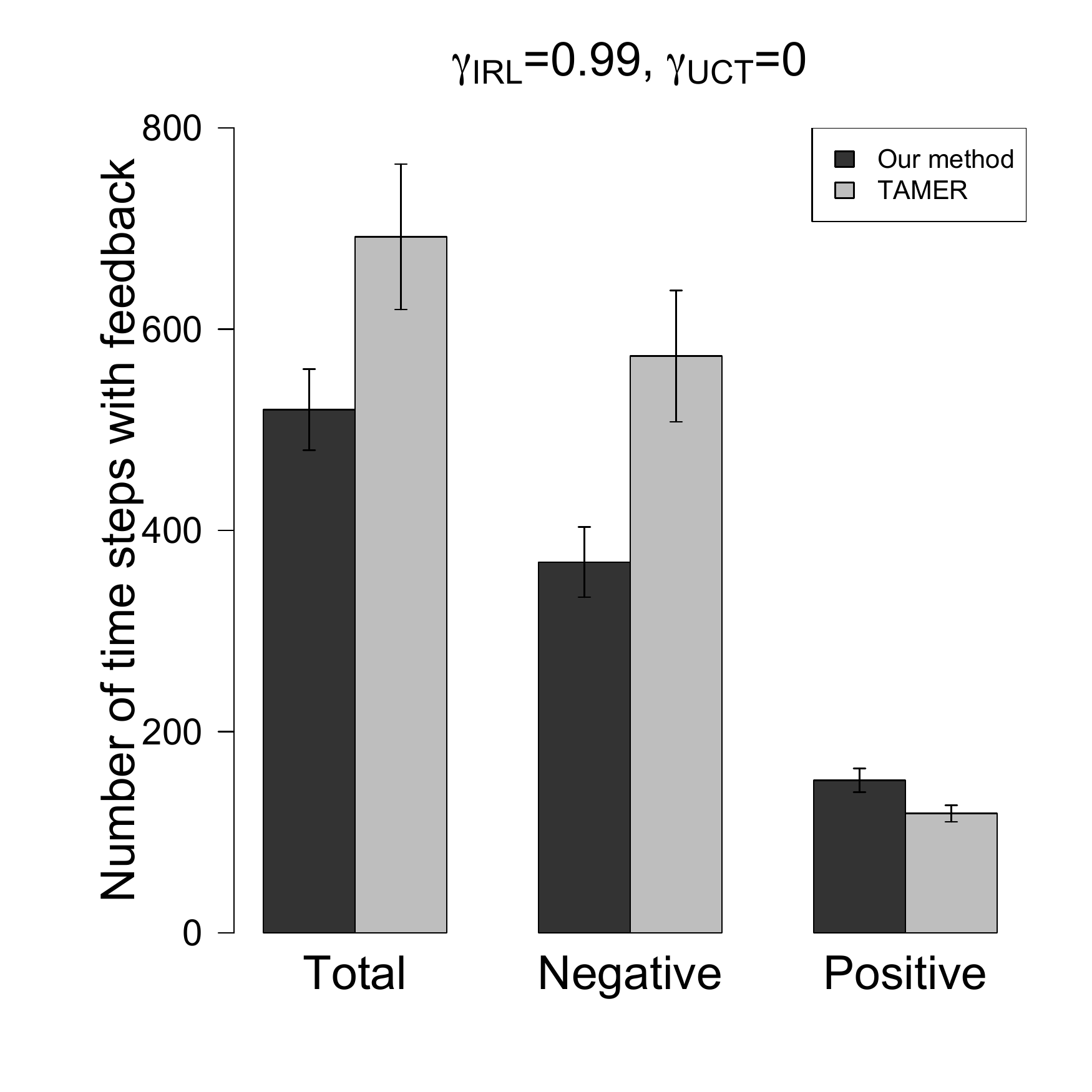}&
\includegraphics[width=0.48\columnwidth,  trim=40 0 0 0, clip]{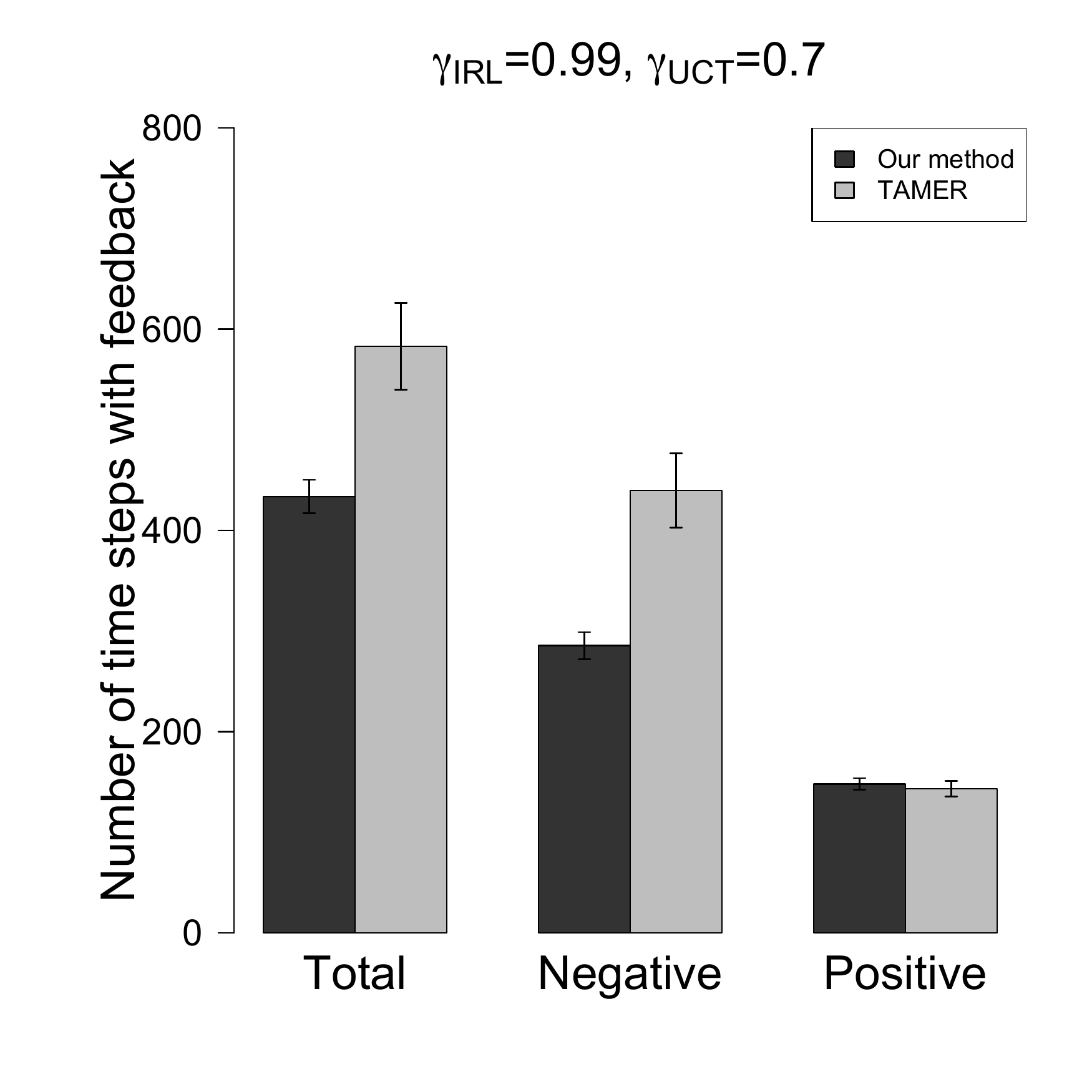}\\
\includegraphics[width=0.48\columnwidth,  trim=40 0 0 0, clip]{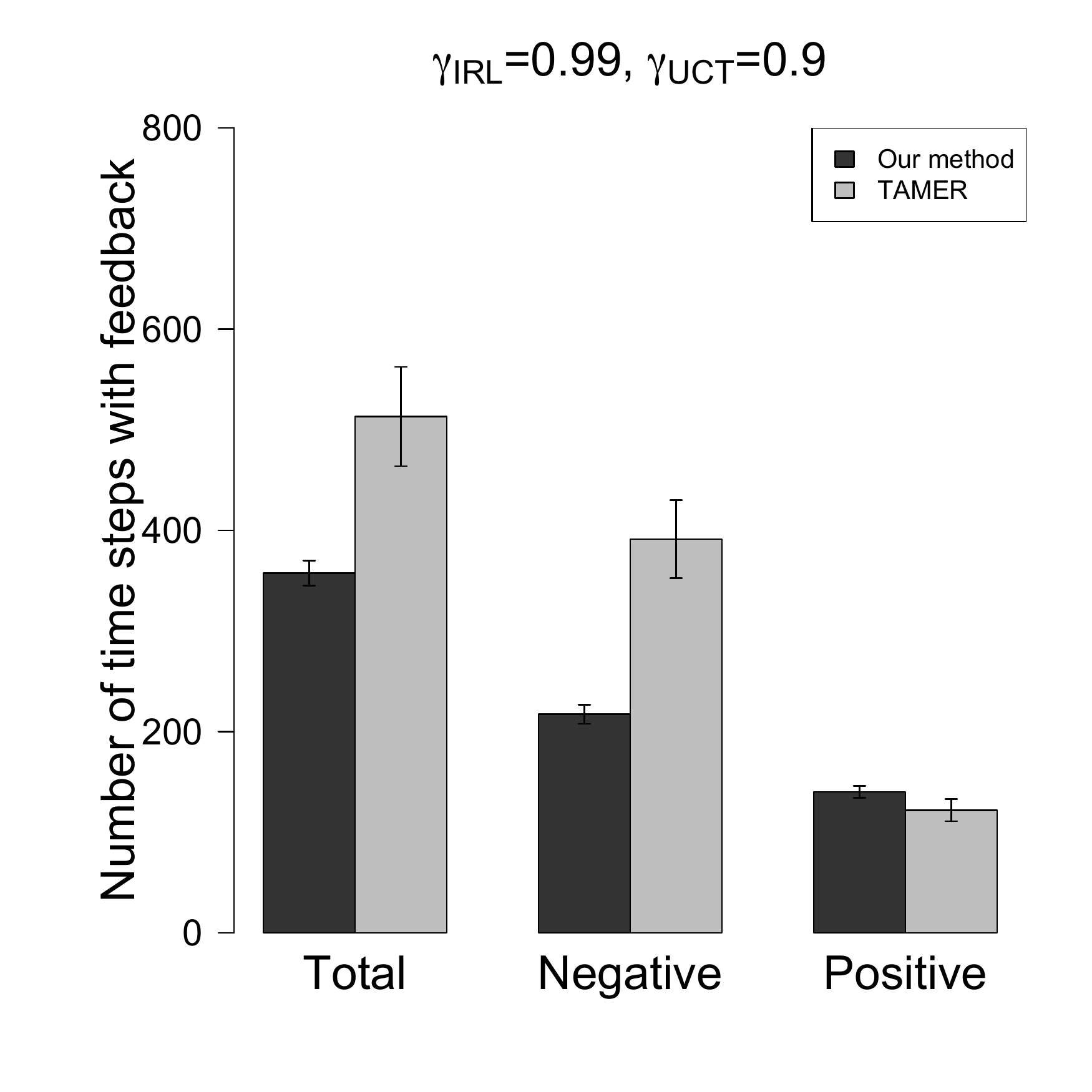}&
\includegraphics[width=0.48\columnwidth,  trim=40 0 0 0, clip]{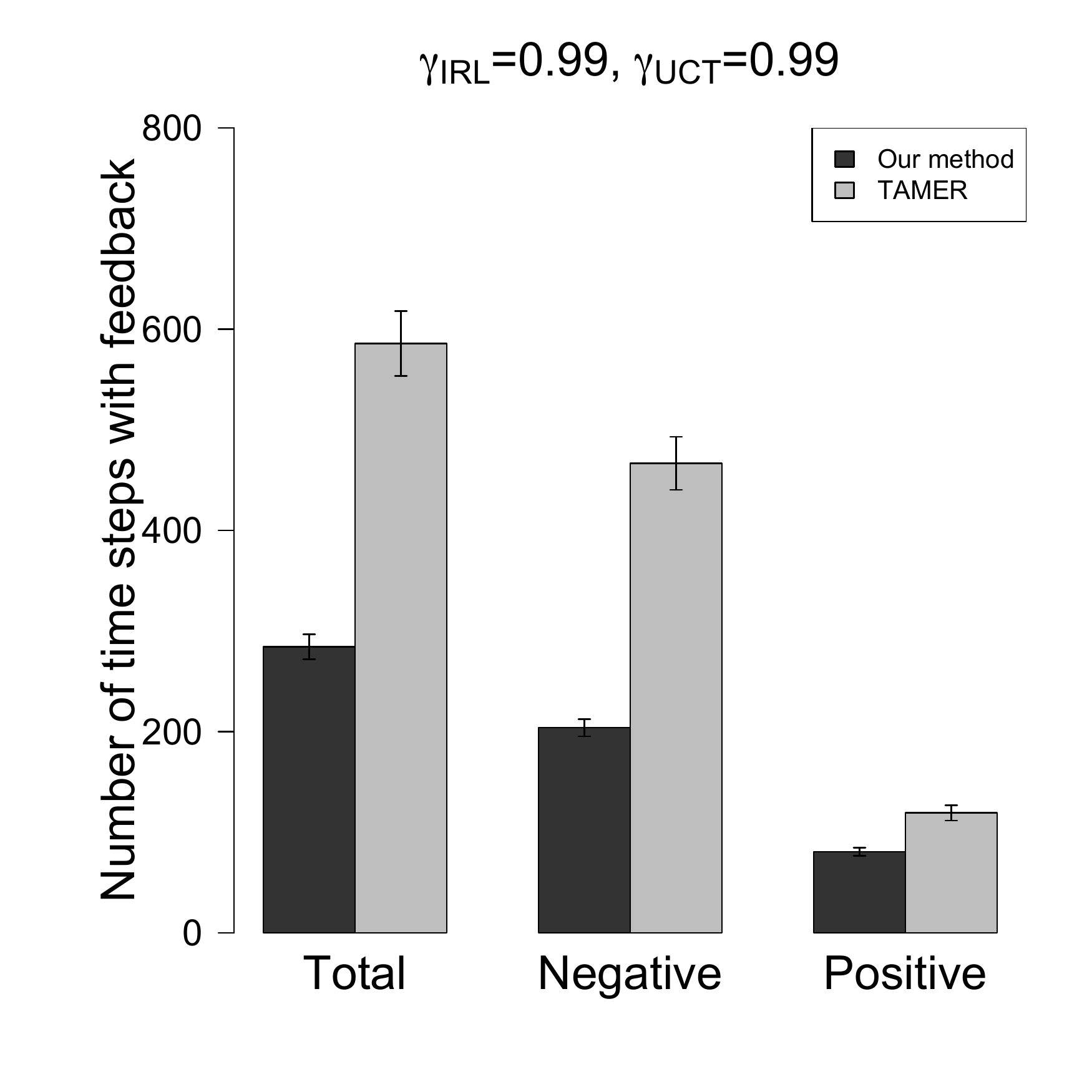} \\
\end{tabular}
\vspace{-4mm}
\caption{Here we can see the number of time steps with feedback trained until an optimal policy is obtained with different discount rates on human reward, in terms of total, positive and negative feedback. Seeding the TAMER agent learning and planning with learned reward function from demonstration can significantly reduce the number of feedback needed to learn an optimal policy. Note: black bars stand for the standard error of the mean.}
\label{feedback}
\end{figure}

We hypothesized that agents trained with our proposed method 
will require less feedback than those with the TAMER 
framework, especially the negative one. 
To measure the amount of feedback given, we counted the number of time steps with feedback, comparing our method 
to TAMER. 
Figure \ref{feedback} shows the number of time step with feedback for both our proposed method 
and TAMER 
agents with different discount rates on human reward, 
in terms of total feedback, positive and negative feedback. 

\begin{table}[t!]
\begin{center}
\caption{Ratio of positive and negative feedback among the total number of feedback given for agent learning with our method and TAMER. 
Note: p-value was computed with t-test, $\sigma$ is the standard deviation.}
\label{table:ratio}
\resizebox{0.95\columnwidth}{!}{ \footnotesize
\begin{tabular}{cccccc}
\toprule[.8pt]
 \multirow{2}{*}{}&&\multicolumn{4}{c}{Discount on human reward ($\gamma_{UCT}$)}\\
 \cline{3-6}
   &&0&0.7&0.9&0.99\\
\midrule[.8pt]
\multirow{2}{*}{
Our method 
} 
&Positive                     &0.30    &0.34    &0.39	&0.28\\
&Negative            &0.70    &0.66    &0.61 	&0.72\\
&                          &($\sigma =0.06$) &($\sigma =0.03$) &($\sigma =0.11$) &($\sigma =0.02$) \\
\midrule[.8pt]
\multirow{2}{*}{
TAMER 
} 
&Positive                     &0.18    &0.25    &0.24	&0.20\\
&Negative            &0.82    &0.75    &0.76 	&0.80\\
&                          &($\sigma =0.03$)  &($\sigma =0.03$) &($\sigma =0.07$) &($\sigma =0.02$) \\
\midrule[.8pt]
\multicolumn{2}{c}{$p-value$ (t-test)}          &6.6e-05           &1.4e-06           &6.4e-10          &6.2e-07       \\
\bottomrule[.8pt]
\end{tabular}
}
\end{center}
\vspace{-2mm}
\end{table}


From Figure \ref{feedback} we can see that, agent learning with our method 
received significantly less total feedback than the TAMER 
agent for all discounts on human reward ($\gamma_{UCT}$ = 0, 0.7, 0.9 and 0.99 for the TAMER planning  respectively). In terms of negative feedback, 
the agent learning with our method 
also received significantly less feedback than the TAMER 
agent. The largest differences between our method and TAMER in terms of total and negative feedback are achieved when the discount rate on human reward is highest (i.e. $\gamma_{UCT}$ = 0.99). 
Although the number of received positive feedback for our proposed method is not always more than that for the TAMER 
agent, Table \ref{table:ratio} shows the ratio of positive feedback among the total number of feedback given for our proposed method is significantly higher than that for TAMER 
agent learning with all discount rates on human reward. 
These results suggest that learning from demonstration can improve the learning efficiency of a TAMER agent by reducing the total number of human rewards needed to train an agent to get an optimal policy. Moreover, the provided demonstration can reduce the number of incorrect actions and increase the ratio of correct actions during the learning process. 


\subsection{Performance}

\begin{figure}[htb]
\centering
\vspace{-6mm}
\begin{tabular}{c}
\includegraphics[width=0.75\columnwidth,  trim=30 0 0 0, clip]{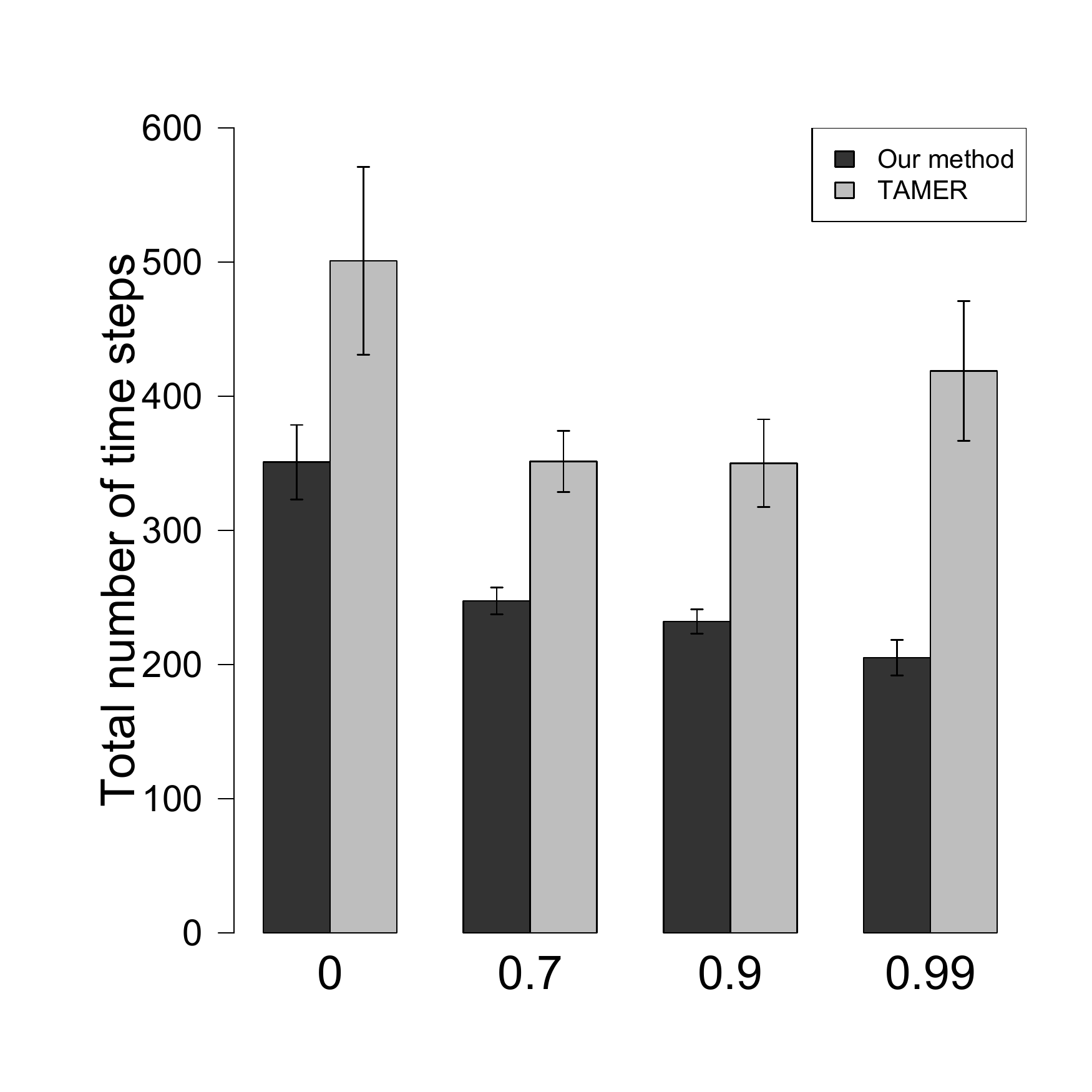}
\end{tabular}
\vspace{-8mm}
\caption{Seeding the TAMER agent learning and planning with learned reward function from demonstration can significantly reduce the total number of time steps needed to obtain an optimal policy with different discount rates on human reward. 
Note: black bars stand for the standard error of the mean.}
\vspace{-2mm}
\label{total}
\end{figure}

Since the task performance metric is based on the time steps taken to reach the goal in the Grid World domain, we take the number of total time steps needed to train the agent to obtain an optimal policy as the performance measure in our experiments. Figure \ref{total} shows the total number of time steps (actions) needed for training an agent to obtain an optimal policy with our proposed method 
and TAMER 
using different discount rates on human reward. From Figure \ref{total} we can see that, the total number of time steps needed to train an agent with our method 
is significantly fewer than a TAMER 
agent for all discount rates on human reward. Figure \ref{total} also shows that the total number of actions needed for an agent to obtain an optimal policy with our proposed method is decreasing when the discount on human reward increasing from 0 to 0.99. The largest difference between our method and TAMER is achieved when the discount factor is 0.99.


\begin{figure}[htb]
\centering
\begin{tabular}{c c}
\includegraphics[width=0.47\columnwidth,  trim=40 10 10 10, clip]{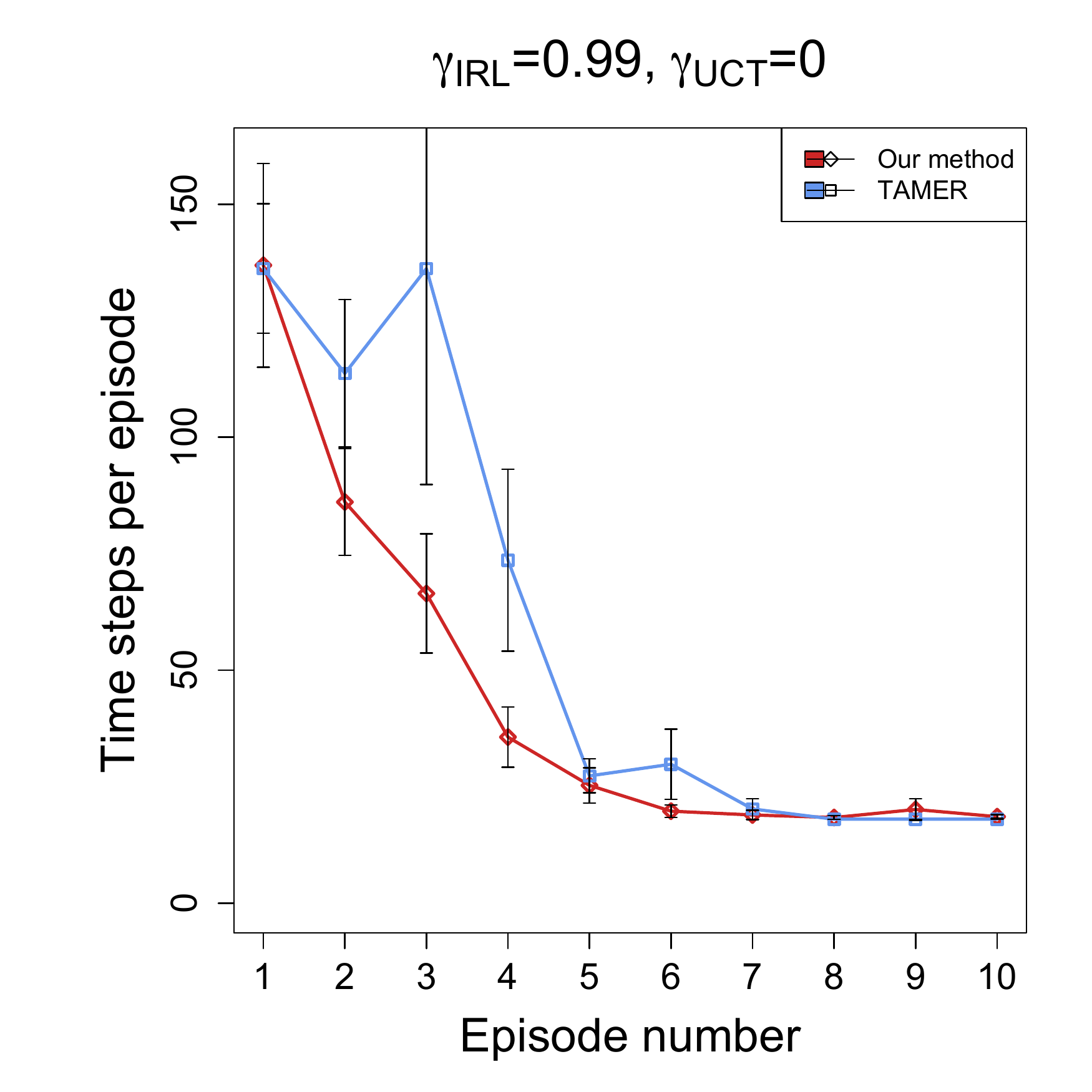}&
\includegraphics[width=0.47\columnwidth,  trim=40 10 10 10, clip]{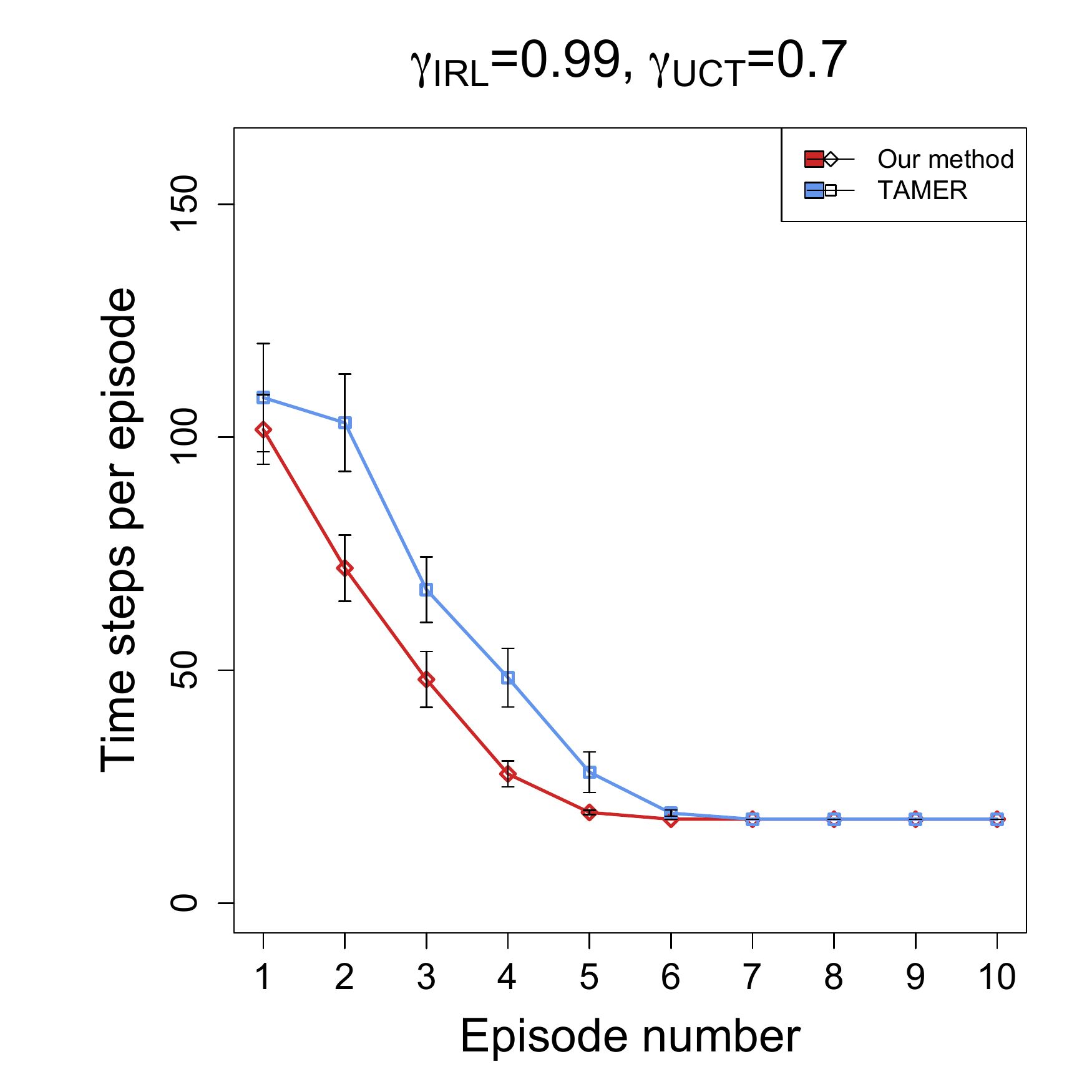}\\
\includegraphics[width=0.47\columnwidth,  trim=40 10 10 10, clip]{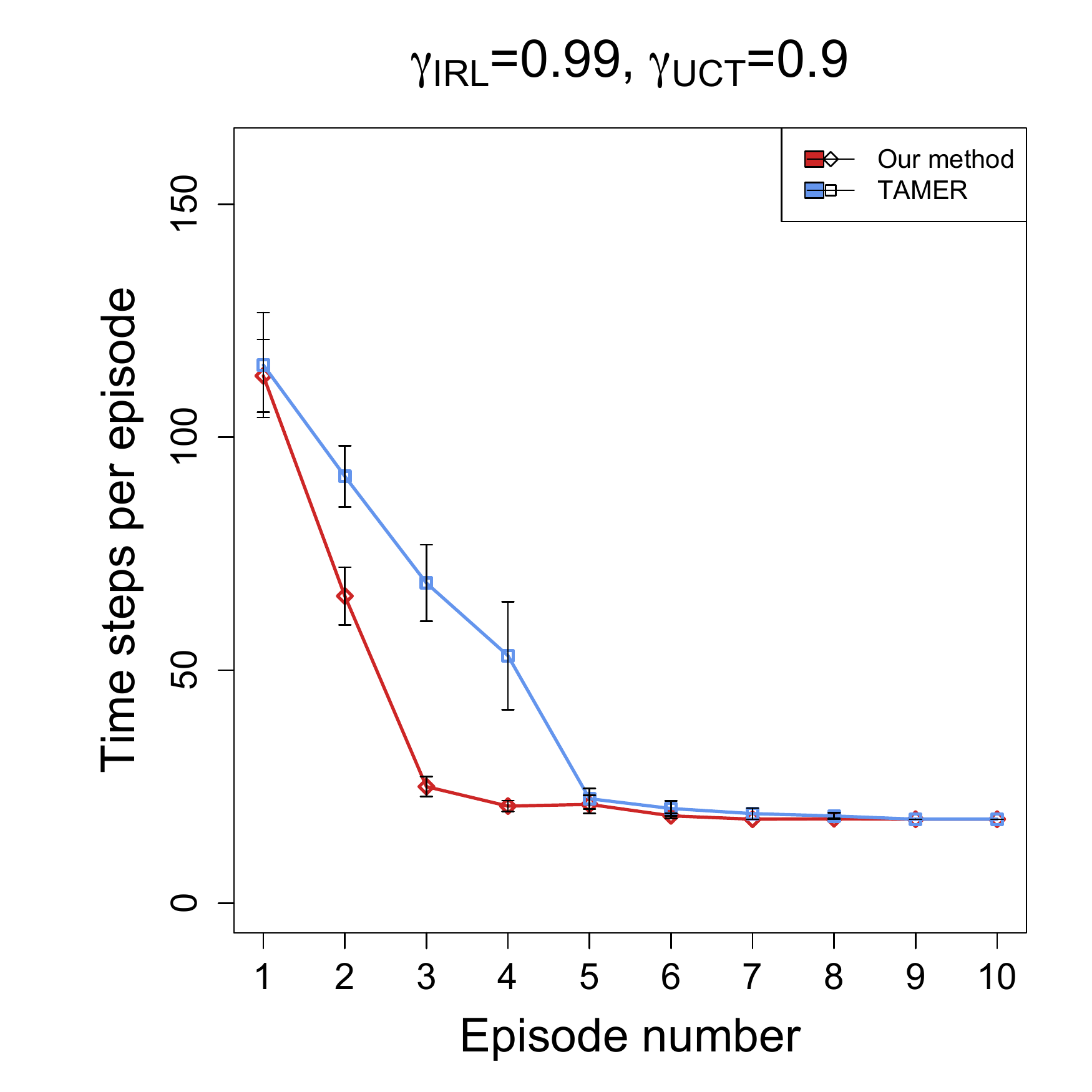}&
\includegraphics[width=0.47\columnwidth,  trim=40 10 10 10, clip]{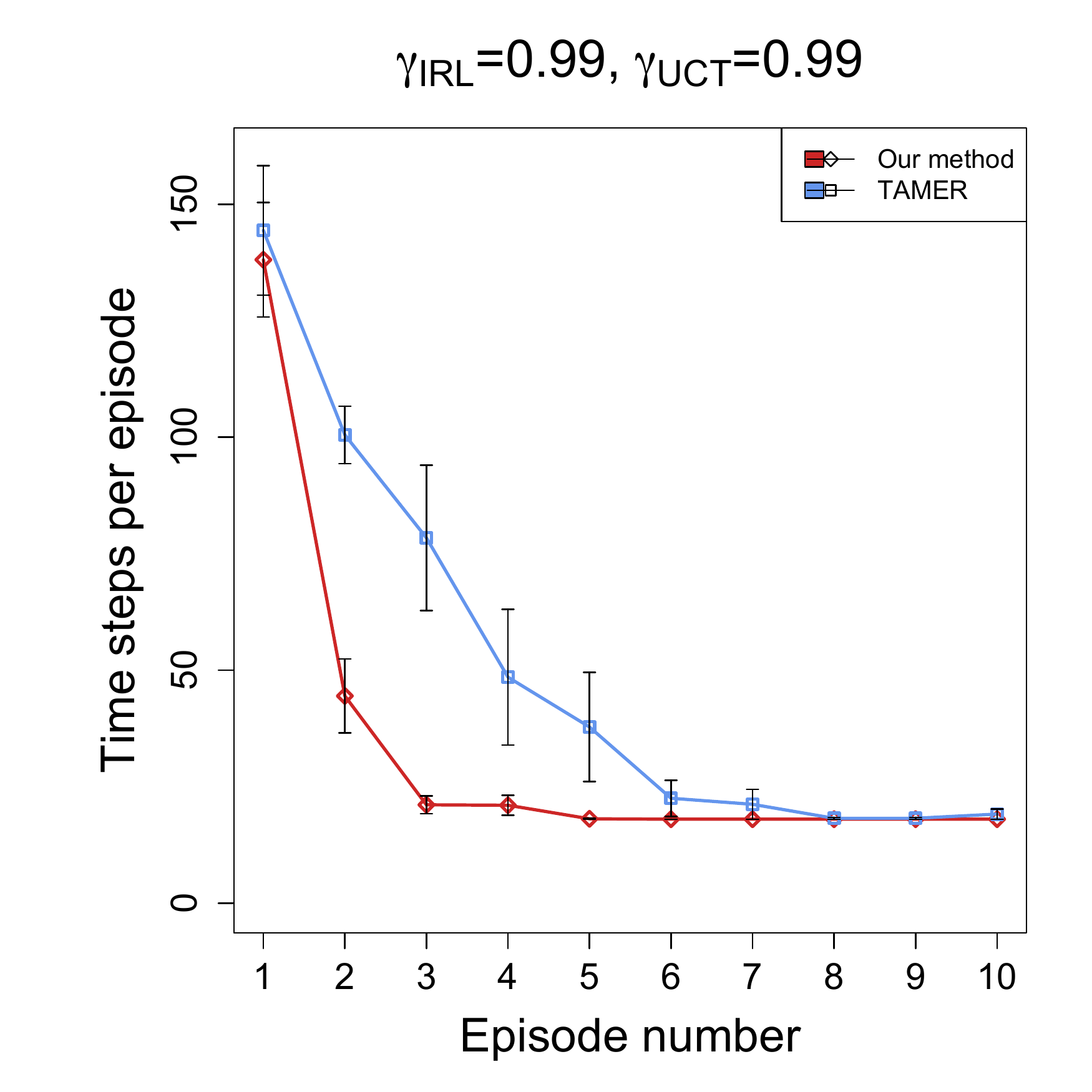} \\
\end{tabular}
\vspace{-2mm}
\caption{Here we can see the number of time steps needed to reach the goal per episode until an optimal policy is obtained with different discount rates on human reward for our proposed method and TAMER. 
Seeding with learned reward function from demonstration allows the TAMER agent learning and planning faster. Note: black bars stand for the standard error of the mean.}
\label{episode}
\end{figure}

We also analyzed the number of time steps per episode trained until an optimal policy is obtained with our proposed method 
and TAMER 
during the training process, as shown in Figure \ref{episode}. From Figure \ref{episode} we can see that, for all discount rates on human reward, 
the number of time steps for the first episode with both our method 
and TAMER 
is similar. 
But after that, the number of time steps for each episode with our proposed method  decreased dramatically especially for $\gamma_{UCT}$ = 0.9 and 0.99, which 
is significantly fewer than that with TAMER 
before obtaining an optimal policy. 
This could be because the TAMER agent plans and updates the value function with UCT, it has not update the value function for all states yet. In this case the learned reward function from demonstration has little effect for TAMER agent planning in the initial training process.
Moreover, for all discounts on human reward, agents with our method converge faster than those with TAMER. 
For example, it takes about one episode fewer for agents with our method to learn an optimal policy in comparison with TAMER while the discounts on human reward are 0 and 0.7.  When $\gamma_{UCT}$ = 0.9 and 0.99, agents with our method learn an optimal policy within three or four episodes, while it takes a TAMER agent five to eight episodes to achieve the same performance.  
Moreover, the number of episodes needed to train an agent with our method to obtain an optimal policy is decreasing 
when the discount on human reward increasing from 0 to 0.99. 

\vspace{-0.4cm}
\section{Discussion}
\label{sec:dis}

\begin{figure} [htb]
\centering
\begin{tabular}{c c}
\includegraphics[width=0.47\columnwidth]{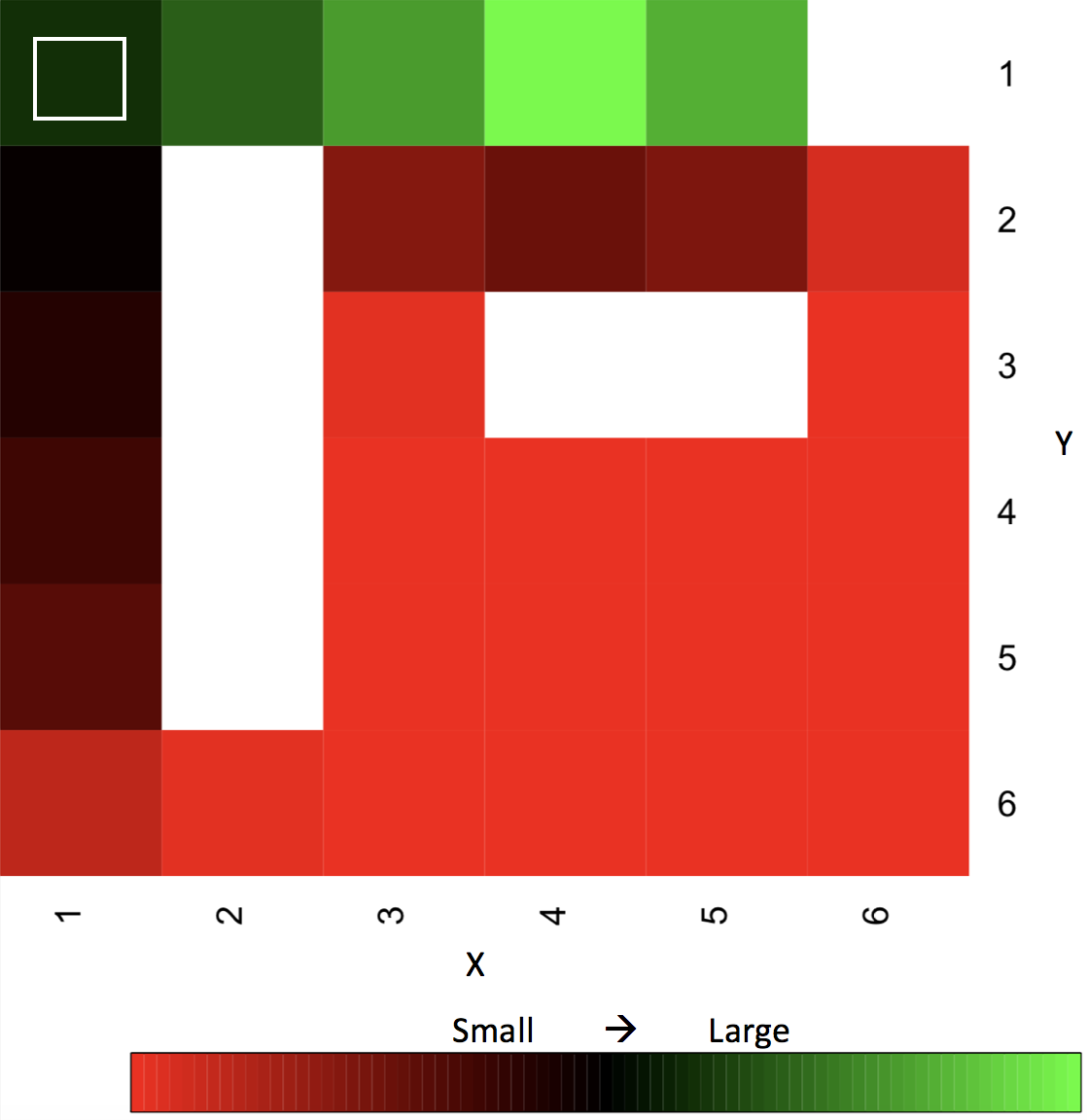}&
\includegraphics[width=0.47\columnwidth]{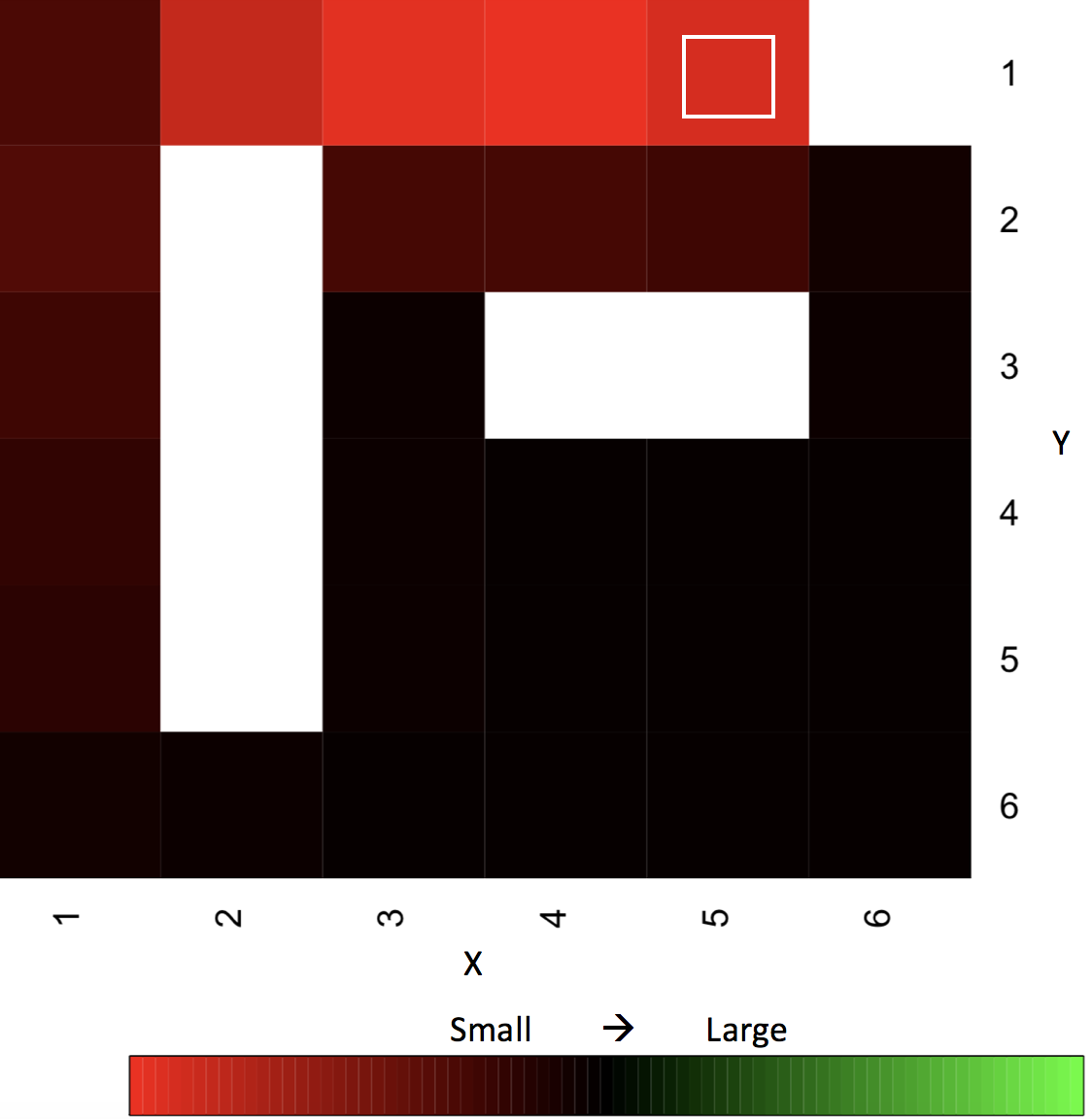}\\
(a) TAMER &(b) Our method\\
\vspace{-0.6cm}
\end{tabular}
\caption{Heat map of state value when the TAMER agent visits the state block X1Y1 for the first time  (a) and when the agent with our proposed method after learning from demonstration via IRL visits the starting state block X5Y1 for the first time (b). Note: White square shows the block the robot is in, X means the row number and Y means the column number in the board.}
\label{heatmap}
\vspace{-0.2cm}
\end{figure}

For a TAMER agent to search via UCT in complex environments with large state space, when the agent does not visit the states near the goal, the human trainer cannot give feedback for those states. This prevents the agent from learning about what reward might be received along those critical states. Therefore, a TAMER agent planning with UCT cannot learn the true values of states along its optimal path. 
During the learning process, the learned policy might take the agent back to previously experienced states since it already received feedback and updated those states with high values. 
The 
heat map of state value function in Figure \ref{heatmap}(a) support this explanation. From Figure \ref{heatmap}(a) we can see  that experienced states are updated with high value while states that are far from experienced states often have not been updated even once and retain their original state values. 
While with our proposed method, the TAMER agent can get a roughly optimal policy up to the deepest search and encourage it 
to explore along the optimal path while navigating to the goal, as shown in Figure \ref{heatmap}(b).

\vspace{-0.4cm}
\section{CONCLUSIONS}
\label{sec:con}


In this paper, to 
alleviate the encountered local-bias problem when the TAMER agent plans with the UCT algorithm and reduce the agent's cost in the learning process, we propose to drive the TAMER agent's exploration along the optimal path by initializing the agent's reward function via learning from demonstration. We test our proposed method in the RL benchmark testing domain ---Grid World---with different discount rates on human reward.  Our results show that learning from demonstration can improve the agent's learning efficiency by reducing 
feedback needed to obtain an optimal policy. 
In addition, 
demonstration can reduce the learning cost by decreasing the number of incorrect actions and increasing the ratio of correct actions during the learning process. 
More importantly,  demonstration allows a TAMER agent to learn a roughly optimal policy up to the deepest search and encourage the agent to explore along the optimal path, allowing it 
to converge faster. 


In future work, we would like to further test our method in 
complex domains 
and conduct a user study, to see how our method and results generalize to other domains and members of the general public. 
In addition, we prefer to further extend our method by combining with deep learning method to see how learning from demonstration can affect an agent's learning and planning from human reward in complex tasks with high-dimensional state spaces. 
\bibliographystyle{named}

\begin{thebibliography}{}

\bibitem[\protect\citeauthoryear{Abbeel and
  Ng}{2004}]{abbeel2004apprenticeship}
Pieter Abbeel and Andrew~Y Ng.
\newblock Apprenticeship learning via inverse reinforcement learning.
\newblock In {\em Proceedings of the 21st International Conference on Machine
  Learning}, pages 1--8. ACM, 2004.

\bibitem[\protect\citeauthoryear{Argall \bgroup \em et al.\egroup
  }{2007}]{argall2007learning}
Brenna Argall, Brett Browning, and Manuela Veloso.
\newblock Learning by demonstration with critique from a human teacher.
\newblock In {\em Proceedings of the ACM/IEEE international conference on
  Human-robot interaction}, pages 57--64. ACM, 2007.

\bibitem[\protect\citeauthoryear{Argall \bgroup \em et al.\egroup
  }{2009}]{argall2009survey}
Brenna~D Argall, Sonia Chernova, Manuela Veloso, and Brett Browning.
\newblock A survey of robot learning from demonstration.
\newblock {\em Robotics and autonomous systems}, 57(5):469--483, 2009.

\bibitem[\protect\citeauthoryear{Brys \bgroup \em et al.\egroup
  }{2015}]{brys2015reinforcement}
Tim Brys, Anna Harutyunyan, Halit~Bener Suay, Sonia Chernova, Matthew~E Taylor,
  and Ann Now{\'e}.
\newblock Reinforcement learning from demonstration through shaping.
\newblock In {\em Proceedings of the International Joint Conference on
  Artificial Intelligence (IJCAI)}, 2015.

\bibitem[\protect\citeauthoryear{Isbell \bgroup \em et al.\egroup
  }{2001}]{isbell2001social}
Charles Isbell, Christian~R Shelton, Michael Kearns, Satinder Singh, and Peter
  Stone.
\newblock A social reinforcement learning agent.
\newblock In {\em Proceedings of the 5th International Conference on Autonomous
  Agents}, pages 377--384. ACM, 2001.

\bibitem[\protect\citeauthoryear{Judah \bgroup \em et al.\egroup
  }{2014}]{judah2014imitation}
Kshitij Judah, Alan Fern, Prasad Tadepalli, and Robby Goetschalckx.
\newblock Imitation learning with demonstrations and shaping rewards.
\newblock In {\em Proceedings of the 28th AAAI Conference on Artificial
  Intelligence}, pages 1890--1896, 2014.

\bibitem[\protect\citeauthoryear{Knox and Stone}{2009}]{knox2009interactively}
W~Bradley Knox and Peter Stone.
\newblock Interactively shaping agents via human reinforcement: the {TAMER}
  framework.
\newblock In {\em Proceedings of the 5th International Conference on Knowledge
  Capture}, pages 9--16. ACM, 2009.

\bibitem[\protect\citeauthoryear{Knox and Stone}{2015}]{knox2015framing}
W~Bradley Knox and Peter Stone.
\newblock Framing reinforcement learning from human reward: reward positivity,
  temporal discounting, episodicity, and performance.
\newblock {\em Artificial Intelligence}, 225:24--50, 2015.

\bibitem[\protect\citeauthoryear{Kocsis and
  Szepesv{\'a}ri}{2006}]{kocsis2006bandit}
Levente Kocsis and Csaba Szepesv{\'a}ri.
\newblock Bandit based monte-carlo planning.
\newblock In {\em European conference on machine learning}, pages 282--293.
  Springer, 2006.

\bibitem[\protect\citeauthoryear{Loftin \bgroup \em et al.\egroup
  }{2014}]{loftin2014strategy}
Robert Loftin, James MacGlashan, M~Littman, M~Taylor, and D~Roberts.
\newblock A strategy-aware technique for learning behaviors from discrete human
  feedback.
\newblock In {\em Proceedings of the 28th AAAI Conference on Artificial
  Intelligence (AAAI-2014)}, 2014.

\bibitem[\protect\citeauthoryear{Loftin \bgroup \em et al.\egroup
  }{2015}]{loftin2015learning}
Robert Loftin, Bei Peng, James MacGlashan, Michael~L Littman, Matthew~E Taylor,
  Jeff Huang, and David~L Roberts.
\newblock Learning behaviors via human-delivered discrete feedback: modeling
  implicit feedback strategies to speed up learning.
\newblock {\em Autonomous Agents and Multi-Agent Systems}, pages 1--30, 2015.

\bibitem[\protect\citeauthoryear{MacGlashan \bgroup \em et al.\egroup
  }{2017}]{macglashan2017interactive}
James MacGlashan, Mark~K Ho, Robert Loftin, Bei Peng, David Roberts, Matthew~E
  Taylor, and Michael~L Littman.
\newblock Interactive learning from policy-dependent human feedback.
\newblock In {\em Proceedings of the 34th International Conference on Machine
  Learning}, volume~70, pages 2285--2294, 2017.

\bibitem[\protect\citeauthoryear{Ng \bgroup \em et al.\egroup
  }{2000}]{ng2000algorithms}
Andrew~Y Ng, Stuart~J Russell, et~al.
\newblock Algorithms for inverse reinforcement learning.
\newblock In {\em Proceedings of International Conference on Machine Learning
  (ICML)}, pages 663--670, 2000.

\bibitem[\protect\citeauthoryear{Pilarski \bgroup \em et al.\egroup
  }{2011}]{pilarski2011online}
Patrick~M Pilarski, Michael~R Dawson, Thomas Degris, Farbod Fahimi, Jason~P
  Carey, and Richard~S Sutton.
\newblock Online human training of a myoelectric prosthesis controller via
  actor-critic reinforcement learning.
\newblock In {\em Proceedings of 12th International Conference on
  Rehabilitation Robotics (ICORR)}, pages 1--7. IEEE, 2011.

\bibitem[\protect\citeauthoryear{Silver \bgroup \em et al.\egroup
  }{2017}]{silver2017mastering}
David Silver, Julian Schrittwieser, Karen Simonyan, Ioannis Antonoglou, Aja
  Huang, Arthur Guez, Thomas Hubert, Lucas Baker, Matthew Lai, Adrian Bolton,
  et~al.
\newblock Mastering the game of go without human knowledge.
\newblock {\em Nature}, 550(7676):354, 2017.

\bibitem[\protect\citeauthoryear{Suay and Chernova}{2011}]{suay2011effect}
Halit~Bener Suay and Sonia Chernova.
\newblock Effect of human guidance and state space size on interactive
  reinforcement learning.
\newblock In {\em Proceedings of IEEE International Symposium on Robot and
  Human Interactive Communication (RO-MAN)}, pages 1--6. IEEE, 2011.

\bibitem[\protect\citeauthoryear{Sutton and
  Barto}{1998}]{sutton1998reinforcement}
R.~Sutton and A.~Barto.
\newblock {\em Reinforcement learning: an introduction}.
\newblock MIT press, 1998.

\bibitem[\protect\citeauthoryear{Thomaz and
  Breazeal}{2008}]{thomaz2008teachable}
Andrea~L Thomaz and Cynthia Breazeal.
\newblock Teachable robots: understanding human teaching behavior to build more
  effective robot learners.
\newblock {\em Artificial Intelligence}, 172(6):716--737, 2008.

\bibitem[\protect\citeauthoryear{Warnell \bgroup \em et al.\egroup
  }{2017}]{warnell2017deep}
Garrett Warnell, Nicholas Waytowich, Vernon Lawhern, and Peter Stone.
\newblock Deep tamer: Interactive agent shaping in high-dimensional state
  spaces.
\newblock {\em arXiv preprint arXiv:1709.10163}, 2017.

\bibitem[\protect\citeauthoryear{Watkins and Dayan}{1992}]{watkins1992q}
Christopher~JCH Watkins and Peter Dayan.
\newblock Q-learning.
\newblock {\em Machine learning}, 8(3-4):279--292, 1992.

\end{thebibliography}

\end{document}